\documentclass[runningheads]{llncs}
\usepackage{graphicx}

\usepackage{tikz}
\usepackage{comment}
\usepackage{amsmath,amssymb, mathtools} 

\usepackage{multirow}
\usepackage{makecell}
\usepackage{color}
\usepackage[export]{adjustbox}
\usepackage{setspace}
\usepackage{booktabs}
\usepackage{tabularx}
\usepackage{array}
\usepackage{subfig}

\usepackage{hyperref}
\hypersetup{pagebackref=true,breaklinks=true,colorlinks,bookmarks=false,citecolor=blue,linkcolor=blue}
\newcommand{\best}[1]{\textcolor{red}{\textbf{#1}}}
\newcommand{\secondbest}[1]{\textcolor{blue}{\underline{#1}}}

\begin{document}
\pagestyle{headings}
\mainmatter
\def\ECCVSubNumber{50}  

\title{AIM 2020 Challenge on \\ Video Temporal Super-Resolution} 

\newcommand {\sanghyun}[1]{{\color{blue}\textbf{Sanghyun: }#1}\normalfont}
\newcommand {\jaerin}[1]{{\color{BurntOrange}\textbf{Jaerin: }#1}\normalfont}
\newcommand {\seungjun}[1]{{\color{MidnightBlue}\textbf{Seungjun: }#1}\normalfont}
\newcommand {\radu}[1]{{\color{ProcessBlue}\textbf{Radu: }#1}\normalfont}
\newcommand {\kyoungmu}[1]{{\color{OrangeRed}\textbf{Kyoung Mu: }#1}\normalfont}


\newcommand{\Paragraph}[1]{\noindent\textbf{#1}}
\newcommand{\etal}{\textit{et al}.}

\titlerunning{AIM 2020 Challenge on Video Temporal Super-Resolution}
%

\author{
    Sanghyun Son\inst{1} \and Jaerin Lee\inst{1} \and Seungjun Nah\inst{1} \and Radu Timofte\inst{2} \and Kyoung Mu Lee\inst{1}
    \and Yihao Liu \and Liangbin Xie \and Li Siyao \and Wenxiu Sun \and Yu Qiao \and Chao Dong 
    \and Woonsung Park \and Wonyong Seo \and Munchurl Kim 
    \and Wenhao Zhang \and Pablo Navarrete Michelini 
    \and Kazutoshi Akita \and Norimichi Ukita 
}

%
\authorrunning{S. Son et al.}
%
\institute{
    CVLab, Seoul National University, Korea \and
    Computer Vision Lab, ETH Z{\"u}rich, Switzerland
}


\maketitle
\begin{abstract}
    Videos in the real-world contain various dynamics and motions that may look unnaturally discontinuous in time when the recorded frame rate is low.
    This paper reports the second AIM challenge on Video Temporal Super-Resolution (VTSR), a.k.a. frame interpolation, with a focus on the proposed solutions, results, and analysis.
    From low-frame-rate (15 fps) videos, the challenge participants are required to submit higher-frame-rate (30 and 60 fps) sequences by estimating temporally intermediate frames.
    To simulate realistic and challenging dynamics in the real-world, we employ the REDS\_VTSR dataset derived from diverse videos captured in a hand-held camera for training and evaluation purposes.
    There have been 68 registered participants in the competition, and 5 teams (one withdrawn) have competed in the final testing phase.
    The winning team proposes the enhanced quadratic video interpolation method and achieves state-of-the-art on the VTSR task.
    %
\end{abstract}

\section{Introduction}
\label{sec:introduction}
{\let\thefootnote\relax\footnotetext{S. Son (sonsang35@gmail.com, Seoul National University), J. Lee, S. Nah, R. Timofte, K. M. Lee are the challenge organizers, while the other authors participated in the challenge.
Appendix~\ref{sec:appendix} contains the authors' teams and affiliations. \\
AIM 2020 webpage:~\url{https://data.vision.ee.ethz.ch/cvl/aim20/}
\\
Competition webpage:~\url{https://competitions.codalab.org/competitions/24584}}}

The growth of broadcasting and streaming services have brought needs for high-fidelity videos in terms of both spatial and temporal resolution.
Despite the advances in modern video technologies capable of playing videos in high frame-rates, e.g., 60 or 120 fps, videos tend to be recorded in lower frame-rates due to the recording cost and quality adversaries~\cite{Nah_2019_CVPR_Workshops_REDS}.
First, fast hardware components are required to process and save large numbers of pixels in real-time.
%
For example, memory and storage should be fast enough to handle at least 60 FHD (1920 $\times$ 1080) video frames per second, while the central processor has to encode those images into a compressed sequence.
Meanwhile, relatively high compression ratios are required to save massive amounts of pixels into slow storage.
Line-skipping may occur at the sensor readout time to compensate for the slow processor.
Also, small camera sensors in mobile devices may suffer from high noise due to the short exposure time as the required frame rate goes higher.
Such limitations can degrade the quality of recorded videos and make them difficult to be acquired in practice.

To overcome those constraints, video temporal super-resolution (VTSR), or video frame interpolation (VFI), algorithms aim to enhance the temporal smoothness of a given video by interpolating missing frames.
For such purpose, several methods~\cite{Jiang_2018_CVPR,Meyer_2015_CVPR,Niklaus_2017_CVPR,Niklaus_2017_ICCV,qvi_nips19,xue2019video} have been proposed to reconstruct smooth and realistic sequences from videos of low-frame-rate, which looks discontinuous in the temporal domain.
Nevertheless, there also exist several challenges in conventional VTSR methods.
First, real-world motions are highly complex and nonlinear, not easy to estimate with simple dynamics.
Also, since the algorithm has to process videos of high-quality and high-resolution, efficient approaches are required to handle thousands of frames swiftly in practice.

To facilitate the development of robust and efficient VTSR methods, we have organized AIM 2020 VTSR challenge.
In this paper, we report the submitted methods and the benchmark results.
Similarly to the last AIM 2019 VTSR challenge~\cite{Nah_2019_ICCV_Workshops_VTSR}, participants are required to recover 30 and 60 fps video from low-frame-rate 15 fps input sequences from the REDS\_VTSR~\cite{Nah_2019_ICCV_Workshops_VTSR} dataset.
5 teams have submitted the final solution, and the XPixel team won the first-place prize with the Enhanced Quadratic Video Interpolation method.
The following sections describe the related works and introduce AIM 2020 Video Temporal Super-Resolution challenge~(VTSR) in detail.
We also present and discuss the challenge results with the proposed approaches.

\section{Related Works}
\label{sec:related}
%
%
While some approaches synthesize intermediate frames directly~\cite{long2016learning,choi2020channel}, most methods embed motion estimation modules into neural network architectures. 
Roughly, the motion modeling could be categorized into phase, kernel, and flow-based approaches. Several methods try to further improve inference accuracy adaptively to the input.

\Paragraph{Phase-based, Kernel-based methods:}
In an early work of Meyer et al.~\cite{Meyer_2015_CVPR}, the temporal change of frames is modeled as phase shifts.
The PhaseNet~\cite{Meyer_2018_CVPR} model has utilized steerable pyramid filters for encoding larger dynamics with a phase shift.
On the other hand, kernel-based approaches use convolutional kernels to shift the input frames to an intermediate position.
The AdaConv~\cite{Niklaus_2017_CVPR} model estimates spatially adaptive kernels, and the SepConv~\cite{Niklaus_2018_CVPR} improves computational efficiency by factorizing the 2D spatial kernel into 1D kernels.

\Paragraph{Flow-based approaches:}
Meanwhile, optical flow-based piece-wise linear models have grown popular.
The DVF~\cite{Liu_2017_ICCV} and SuperSloMo~\cite{Jiang_2018_CVPR} models estimate optical flow between two input frames and warp them to the target intermediate time.
Later, the accuracy of SuperSloMo is improved by training with cycle consistency loss~\cite{Reda_2019_ICCV}.
The TOF~\cite{xue2019video}, CtxSyn~\cite{Niklaus_2018_CVPR}, and BMBC~\cite{Park_2020_ECCV} models additionally apply image synthesis modules on the warped frames to produce the intermediate frame.
To interpolate high-resolution videos in fast speed, the IM-Net~\cite{Peleg_2019_CVPR} uses block-level horizontal/vertical motion vectors in multi-scale.

It is essential to handle the occluded area when warping with the optical flow as multiple pixels could overlap in the warped location.
Jiang~\etal~\cite{Jiang_2018_CVPR} and Yuan~\etal~\cite{Yuan_2019_CVPR} estimate occlusion maps to exclude invalid contributions.
Additional information is explored, such as depth~\cite{Bao_2019_CVPR} and scene context~\cite{Niklaus_2018_CVPR} to determine proper local warping weights.
Niklaus~\etal~\cite{Niklaus_2020_CVPR} suggests forward warping with the softmax splatting method, while many other approaches use backward warping with sampling.
    
Several works tried mixed motion representation of kernel-based and flow-based models.
The DAIN~\cite{Bao_2019_CVPR} and MEMC-Net~\cite{MEMC-Net} models use both the kernel and optical flow to apply adaptive warping.
The AdaCof~\cite{Lee_2020_CVPR} unifies the combined representation via an adaptive collaboration of flows in a similar formulation as deformable convolution~\cite{dai2017deformable}.
To enhance the degree of freedom in the representation, the local weight values differ by position. 
    
Despite the success of motion estimation with optical flow, complex motions in the real world are hard to be modeled due to the simplicity of optical flow.
Most of the methods using the optical flow between the two frames are limited to locally linear models.
To improve the degree of freedom in the motion model, higher-order representations are proposed.
Quadratic~\cite{qvi_nips19,qvi_iccvw19} and cubic~\cite{Chi_2020_ECCV} motion flow is estimated from multiple input frames.
Such higher-order polynomial models could reflect accelerations and nonlinear trajectories.
In AIM 2019 Challenge on Video Temporal Super-Resolution, the quadratic video interpolation method~\cite{qvi_nips19,qvi_iccvw19} achieved the best accuracy. 
    
\Paragraph{Refinement methods:}
On top of the restored frames, some methods try to improve the initial estimation with various techniques.
Gui~\etal~\cite{Gui_2020_CVPR} propose a 2-stage estimation method.
Initial interpolation is obtained with a structure guide in the first stage and the second stage refines the texture.
Choi~\etal~\cite{Choi_2020_CVPR} perform test-time adaptation of the model via meta-learning.
Motivated by classical pyramid energy minimization, Zhang~\etal~\cite{Zhang_2020_ECCV} propose a recurrent residual pyramid network. Residual displacements are refined via recurrence.

\begin{figure}[t]
    \centering
    \subfloat[`\emph{00000064}']{\includegraphics[width=0.195\linewidth]{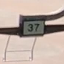}}
    \hfill
    \subfloat[`\emph{00000066}']{\includegraphics[width=0.195\linewidth]{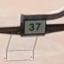}}
    \hfill
    \subfloat[`\emph{00000068}']{\includegraphics[width=0.195\linewidth]{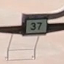}}
    \hfill
    \subfloat[`\emph{00000070}']{\includegraphics[width=0.195\linewidth]{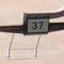}}
    \hfill
    \subfloat[`\emph{00000072}']{\includegraphics[width=0.195\linewidth]{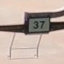}}
    \\
    \caption{
        \textbf{An example of the VTSR task.}
        In the 15 fps $\rightarrow$ 30 fps task, i.e., $\times 2$ VTSR, (c) should be estimated from (a) and (e).
        In the challenging 15 fps $\rightarrow$ 60 fps tasks, i.e., $\times 4$ VTSR, the goal is to predict (b), (c), and (d) using (a) and (e) only.
        We provide a sample sequence from the REDS\_VTSR~\cite{Nah_2019_CVPR_Workshops_REDS} test split.
        Each label indicates the name of the file from a test sequence `\emph{022},' and patches are cropped from the original full-resolution images for better visualization.
        We note that the ground-truth frames (b), (c), and (d) are not provided to participants.
    }
    \label{fig:reds}
\end{figure}

\section{AIM 2020 VTSR Challenge}
This challenge is one of the AIM 2020 associated challenges on:
scene relighting and illumination estimation~\cite{elhelou2020aim_relighting}, image extreme inpainting~\cite{ntavelis2020aim_inpainting}, learned image signal processing pipeline~\cite{ignatov2020aim_ISP}, rendering realistic bokeh~\cite{ignatov2020aim_bokeh}, real image super-resolution~\cite{wei2020aim_realSR}, efficient super-resolution~\cite{zhang2020aim_efficientSR}, and video extreme super-resolution~\cite{fuoli2020aim_VXSR}.
Our development phase has started on May 1st, and the test phase is opened on July 10th.
Participants are required to prepare and submit their final interpolation results in one week.
In AIM 2020 VTSR challenge, there have been a total of 68 participants registered in the CodaLab.
5 teams, including one withdrawn, have submitted their final solution after the test phase.

\label{sec:challenge}
\subsection{Challenge goal}
The purpose of the AIM 2020 VTSR challenge is to develop state-of-the-art VTSR algorithms and benchmark various solutions. 
Participants are required to reconstruct 30 and 60 fps video sequences from 15 fps inputs.
We provide 240 training sequences and 30 validation frames from the REDS\_VTSR~\cite{Nah_2019_CVPR_Workshops_REDS} dataset with 60 fps ground-truth videos.
The REDS\_VTSR dataset is captured by GoPro 6 camera and contains HD-quality (1280 $\times$ 720) videos with 43,200 independent frames for training.
At the end of the test phase, we evaluate the submitted methods on disjoint 30 test sequences. 
Fig.~\ref{fig:reds} shows a sample test sequence.

\subsection{Evaluation metrics}
In conventional image restoration problems, PSNR and SSIM between output and ground-truth images are considered standard evaluation metrics.
In the AIM 2020 VTSR challenge, we use the PSNR as a primary metric, while SSIM values are also provided.
We also require all participants to submit their codes and factsheets at the submission time for the fair competition.
We then check consistency between submitted result images and reproduced outputs by challenge organizers to guarantee reproducible methods.
Since VTSR models are required to process thousands of frames swiftly in practice, one of the essential factors in the algorithms is their efficiency.
Therefore, we measure the runtime of each method in a unified framework to provide a fair comparison.
Finally, this year, we have experimentally included a new perceptual metric, LPIPS~\cite{zhang2018unreasonable}, to assess the visual quality of result images.
Similar to the PSNR and SSIM, the reference-based LPIPS metric is also calculated between interpolated and ground-truth frames.
However, it is known that the LPIPS shows a better correlation with human perception~\cite{zhang2018unreasonable} than the conventional PSNR and SSIM.
We describe more details in Section~\ref{ssec:percep}.

\begin{table}[t]
    \centering
    \caption{
        \textbf{AIM 2020 Video Temporal Super-Resolution Challenge results on the REDS\_VTSR test data.}
        Teams are sorted by PSNR(dB).
        The runtime is measured using a single RTX 2080 Ti GPU based on the submitted codes.
        However, due to the tight memory constraint, we execute the TTI team's method~($^\dagger$) on a Quadro RTX 8000 GPU, which is $\sim$10\% faster than the RTX model.
        SenseSloMo is the winning team from the last AIM 2019 VTSR challenge~\cite{Nah_2019_ICCV_Workshops_VTSR}.
        SepConv~\cite{Niklaus_2017_ICCV} and Baseline overlay methods are also provided for reference.
        The best and second-best are marked with \best{red} and \secondbest{blue}, respectively.
    }
    \label{table_results}
    \begin{tabularx}{\linewidth}{p{5.6cm} >{\centering\arraybackslash}X >{\centering\arraybackslash}X >{\centering\arraybackslash}X >{\centering\arraybackslash}X >{\centering\arraybackslash}X}
        \toprule
        & \multicolumn{2}{c}{15fps $\rightarrow$ 30fps} & \multicolumn{2}{c}{15fps $\rightarrow$ 60fps} & Runtime \\
        Team & PSNR$^\uparrow$ & SSIM$^\uparrow$ & PSNR$^\uparrow$ & SSIM$^\uparrow$ & (s/frame) \\
        \midrule
        \textbf{XPixel~\cite{liu2020enhanced} (Challenge Winner)} & \best{24.78} & \secondbest{0.7118} & \best{25.69} & \secondbest{0.7425} & 12.40 \\
        KAIST-VICLAB  & \secondbest{24.69} & \best{0.7142} & \secondbest{25.61} & \best{0.7462} & 1.57 \\
        BOE-IOT-AIBD~(Reproduced) & 24.49 & 0.7034 & 25.27 & 0.7326 & 1.00 \\
        TTI & 23.59 & 0.6720 & 24.36 & 0.6995 & 6.45$^\dagger$ \\
        \midrule
        BOE-IOT-AIBD~(Submission) & 24.40 & 0.6972 & 25.19 & 0.7269 & - \\
        \textit{Withdrawn team} & 24.29 & 0.6977 & 25.05 & 0.7267 & 2.46\\
        \midrule
        SenseSloMo (AIM 2019 Winner) & 24.56 & 0.7065 & 25.47 & 0.7383 & - \\
        SepConv~\cite{Niklaus_2017_ICCV} & 22.48 & 0.6287 & 23.40 & 0.6631 & - \\
        Baseline~(overlay) & 19.68 & 0.6384 & 20.39 & 0.6625 & - \\
        \bottomrule
    \end{tabularx}
\end{table}

\begin{figure}[t]
    \centering
    \subfloat[GT]{\includegraphics[width=0.195\linewidth]{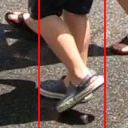}}
    \hfill
    \subfloat[\best{27.04}/\best{0.893}]{\includegraphics[width=0.195\linewidth]{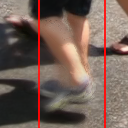}}
    \hfill
    \subfloat[\secondbest{25.67}/\secondbest{0.873}]{\includegraphics[width=0.195\linewidth]{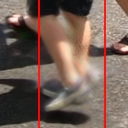}}
    \hfill
    \subfloat[25.55/0.837]{\includegraphics[width=0.195\linewidth]{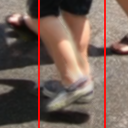}}
    \hfill
    \subfloat[25.68/0.853]{\includegraphics[width=0.195\linewidth]{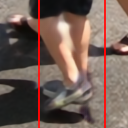}}
    \\
    \caption{
        \textbf{15 $\rightarrow$ 60 fps VTSR results on the REDS\_VTSR dataset.}
        We provide PSNR/SSIM values of each sample for reference.
        (a) is a ground-truth frame, and (b) $\sim$ (e) show result images from the XPixel, KAIST-VICLAB, BOE-IOT-AIBD, and TTI team, respectively.
        \textcolor{red}{Red} lines are drawn to compare alignments with the ground-truth image.
        Each frame is cropped from the test example `\emph{007/00000358}.'
        Best viewed in digital zoom.
    }
    \label{fig:comparison}
\end{figure}

\begin{figure}[t!]
    \centering
    \subfloat[GT]{\includegraphics[width=0.195\linewidth]{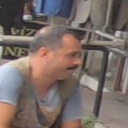}}
    \hfill
    \subfloat[\best{32.64}/\best{0.934}]{\includegraphics[width=0.195\linewidth]{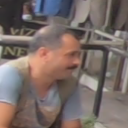}}
    \hfill
    \subfloat[22.42/0.684]{\includegraphics[width=0.195\linewidth]{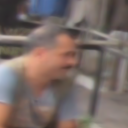}}
    \hfill
    \subfloat[23.86/0.714]{\includegraphics[width=0.195\linewidth]{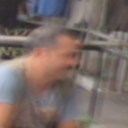}}
    \hfill
    \subfloat[\secondbest{30.64}/\secondbest{0.891}]{\includegraphics[width=0.195\linewidth]{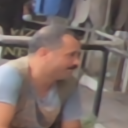}}
    \\
    \subfloat[GT]{\includegraphics[width=0.195\linewidth]{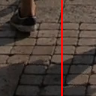}}
    \hfill
    \subfloat[\secondbest{29.33}/\secondbest{0.883}]{\includegraphics[width=0.195\linewidth]{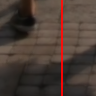}}
    \hfill
    \subfloat[\best{30.19}/\best{0.900}]{\includegraphics[width=0.195\linewidth]{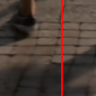}}
    \hfill
    \subfloat[27.86/0.866]{\includegraphics[width=0.195\linewidth]{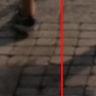}}
    \hfill
    \subfloat[20.30/0.619]{\includegraphics[width=0.195\linewidth]{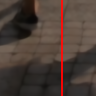}}
    \\
    \subfloat[GT]{\includegraphics[width=0.195\linewidth]{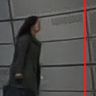}}
    \hfill
    \subfloat[23.32/0.718]{\includegraphics[width=0.195\linewidth]{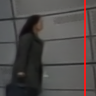}}
    \hfill
    \subfloat[22.92/0.714]{\includegraphics[width=0.195\linewidth]{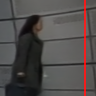}}
    \hfill
    \subfloat[\secondbest{23.85}/\secondbest{0.761}]{\includegraphics[width=0.195\linewidth]{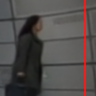}}
    \hfill
    \subfloat[\best{29.11}/\best{0.871}]{\includegraphics[width=0.195\linewidth]{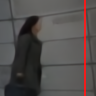}}
    \\
    \caption{
        \textbf{15 $\rightarrow$ 60 fps VTSR results with high performance variance on the REDS\_VTSR dataset.}
        (b) $\sim$ (e), (g) $\sim$ (j), and (l) $\sim$ (o) show result images from the XPixel, KAIST-VICLAB, BOE-IOT-AIBD, and TTI team on the test input `\emph{016/00000358},' `\emph{019/00000182},' and `\emph{029/00000134},' respectively.
    }
    \label{fig:variance}
\end{figure}

\section{Challenge Results}
\label{sec:results}
Table~\ref{table_results} shows the overall summary of the proposed methods from the AIM 2020 VTSR challenge.
The XPixel team has won the first prize, while the KAIST-VICLAB team follows by small margins.
We also provide visual comparisons between the submitted methods.
Fig.~\ref{fig:comparison} illustrates how various algorithms perform differently on a given test sequence.

\subsection{Result analysis}
In this section, we focus on some test examples where the proposed methods show interesting behaviors.
First, we pick several cases where different methods show high performance, i.e., PSNR, variance.
Such analysis can show an advantage of a specific method over the other approaches.
In the first row of Fig.~\ref{fig:variance}, the XPixel team outperforms all the other by a large margin by achieving 32.64dB PSNR.
Compared to the other results, it is clear that an accurate reconstruction of the intermediate frame has contributed to high performance.
On the other hand, the KAIST-VICLAB and TTI team show noticeable performance in the second and third row of Fig.~\ref{fig:variance}, respectively.
In those cases, precise alignments play a critical role in achieving better performance.
Specifically, Fig.~\ref{fig:variance}(i) looks perceptually more pleasing than Fig.~\ref{fig:variance}(h) while there is a significant gap between PSNR and SSIM of those interpolated frames.
Fig.~\ref{fig:variance}(o) shows over +5dB gain compared to all the other approaches.
Although the result frames all look very similar, an accurate alignment algorithm has brought such large gain to the TTI team.

\begin{figure}[t]
    \centering
    \subfloat[GT]{\includegraphics[width=0.195\linewidth]{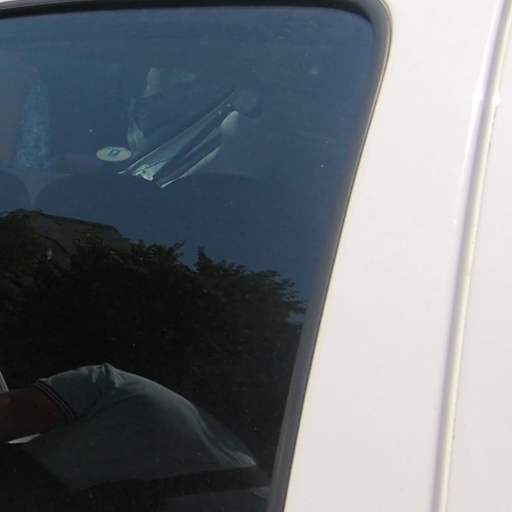}}
    \hfill
    \subfloat[\best{23.90}/\best{0.775}]{\includegraphics[width=0.195\linewidth]{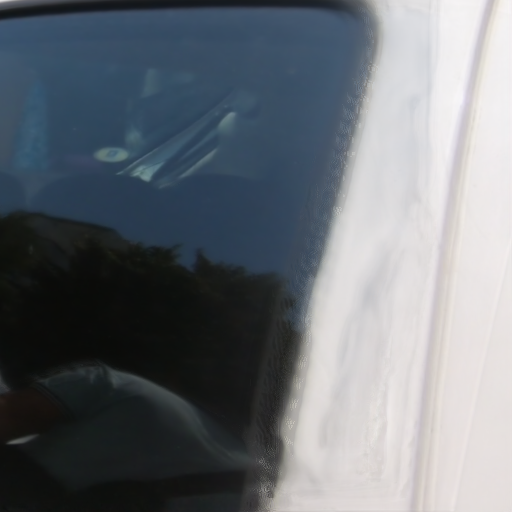}}
    \hfill
    \subfloat[\secondbest{21.80}/0.736]{\includegraphics[width=0.195\linewidth]{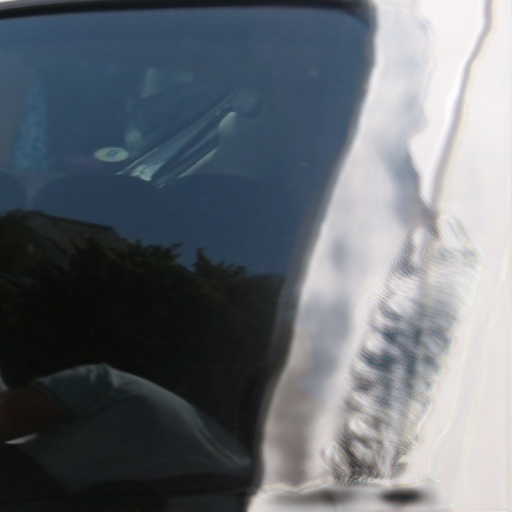}}
    \hfill
    \subfloat[20.96/\secondbest{0.753}]{\includegraphics[width=0.195\linewidth]{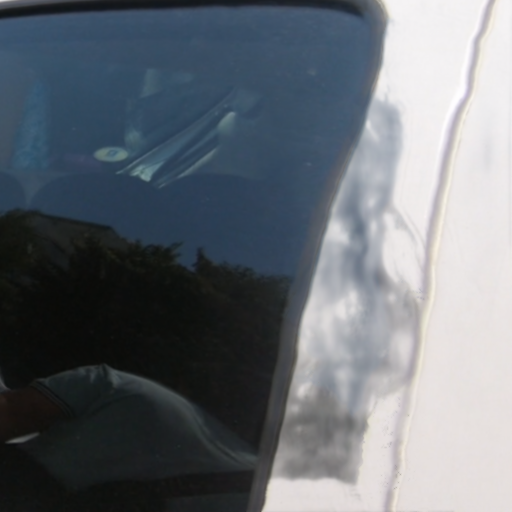}}
    \hfill
    \subfloat[18.72/0.692]{\includegraphics[width=0.195\linewidth]{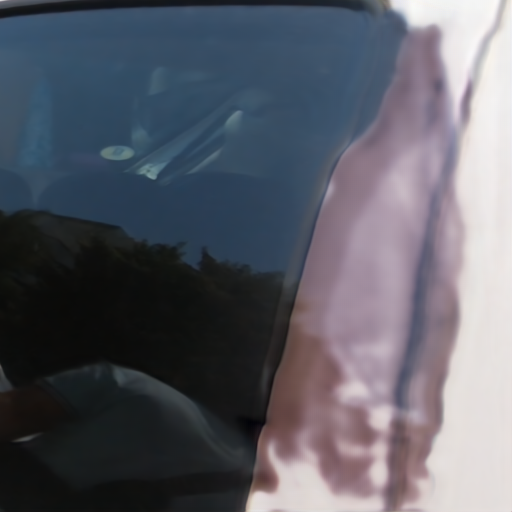}}
    \\
    \caption{
        \textbf{Failure cases of 15 $\rightarrow$ 60 fps VTSR on the REDS\_VTSR dataset.}
        (a) is a ground-truth frame, and (b) $\sim$ (e) show result images from the XPixel, KAIST-VICLAB, BOE-IOT-AIBD, and TTI team, respectively.
        Each frame is cropped from the test example `\emph{021/00000218}.'
    }
    \label{fig:failure}
\end{figure}

In Fig.~\ref{fig:failure}, we also introduce a case where all of the proposed methods are not able to generate clean output frames.
While the submitted models can handle edges and local structures to some extent, they cannot deal with large plain regions and generate unpleasing artifacts.
Such limitation shows the difficulty of the VTSR task on real-world videos and implies several rooms to be improved in the following research.

\subsection{Perceptual quality of interpolated frames}
\label{ssec:percep}
Recently, there have been rising needs for considering the visual quality of output frames in image restoration tasks.
Therefore, we have experimentally included the LPIPS~\cite{zhang2018unreasonable} metric to evaluate how the result images are perceptually similar to the ground-truth.
Table~\ref{tab:lpips} compares the LPIPS score of the submitted methods.
The KAIST-VICLAB team has achieved the best~(lowest) LPIPS on both 15 fps $\rightarrow$ 30 fps and 15 fps $\rightarrow$ 60 fps VTSR tasks even they show lower PSNR compared to the XPixel team.
%
However, we have also observed that the perceptual metric has a weakness to be applied to video-related tasks directly.
For example, the result from KAIST-VICLAB team in Fig.~\ref{fig:variance}(m) has an LPIPS of 0.100, while the TTI team in Fig.~\ref{fig:variance}(o) shows an LPIPS of 0.158.
As the score shows, the LPIPS metric does not consider whether the output frame is accurately aligned with the ground-truth.
In other words, an interpolated sequence with a better perceptual score may look less natural, regardless of how each frame looks realistic.
In future challenges, it would be interesting to develop a video-specific perceptual metric to overcome the limitation of the LPIPS.

\begin{table}[t]
    \centering
    \caption{
        \textbf{LPIPS$_\downarrow$ of the submitted methods on the REDS\_VTSR dataset.}
        All values are calculated from reproduced results.
    }
    \label{tab:lpips}
    \begin{tabularx}{\linewidth}{p{2.5cm} >{\centering\arraybackslash}X >{\centering\arraybackslash}p{3cm} >{\centering\arraybackslash}p{3cm} >{\centering\arraybackslash}X}
        \toprule
        Task $\backslash$ Team & XPixel & KAIST-VICLAB & BOE-IOT-AIBD & TTI \\
        \midrule
        15 fps $\rightarrow$ 30 fps & 0.268 & \best{0.222} & \secondbest{0.249} & 0.289 \\
        15 fps $\rightarrow$ 60 fps & \secondbest{0.214} & \best{0.181} & 0.230 & 0.253 \\
        \bottomrule
    \end{tabularx}
\end{table}

\section{Challenge Methods and Teams}
\label{sec:methods}
This section briefly describes each of the submitted methods in the AIM 2020 VTSR challenge.
Teams are sorted by their final ranking.
Interestingly, the top three teams (XPixel, KAIST-VICLAB, and BOE-IOT-AIBD) have leveraged the QVI model as a baseline and improved the method by their novel approaches.
We briefly describe each method base on the submitted factsheets.
%

\subsection{XPixel}
\Paragraph{Method.}
The XPixel team has proposed the Enhanced Quadratic Video Interpolation~(EQVI)~\cite{liu2020enhanced} method.
The algorithm is built upon the AIM 2019 VTSR Challenge winning approach, QVI~\cite{qvi_nips19,qvi_iccvw19}, with three main components: rectified quadratic flow prediction~(RQFP), residual contextual synthesis network~(RCSN), and multi-scale fusion network~(MS-Fusion).
%
%
To ease the overall optimization procedure and verify the performance gain of each component, four-stage training, and fine-tuning strategies are adopted.
First, the baseline QVI model is trained with a ScopeFlow~\cite{scope} flow estimation method instead of the PWC-Net~\cite{sun2018pwc}.
%
Then, the model is fine-tuned with an additional residual contextual synthesis network.
In the third stage, the rectified quadratic flow prediction is adopted to improve the model with fine-tuning all the modules except the optical flow estimation network.
Finally, all the modules are assembled into the multi-scale fusion network and fine-tuned.
%

%
\begin{figure}[t!]
    \centering
    \includegraphics[width=\linewidth]{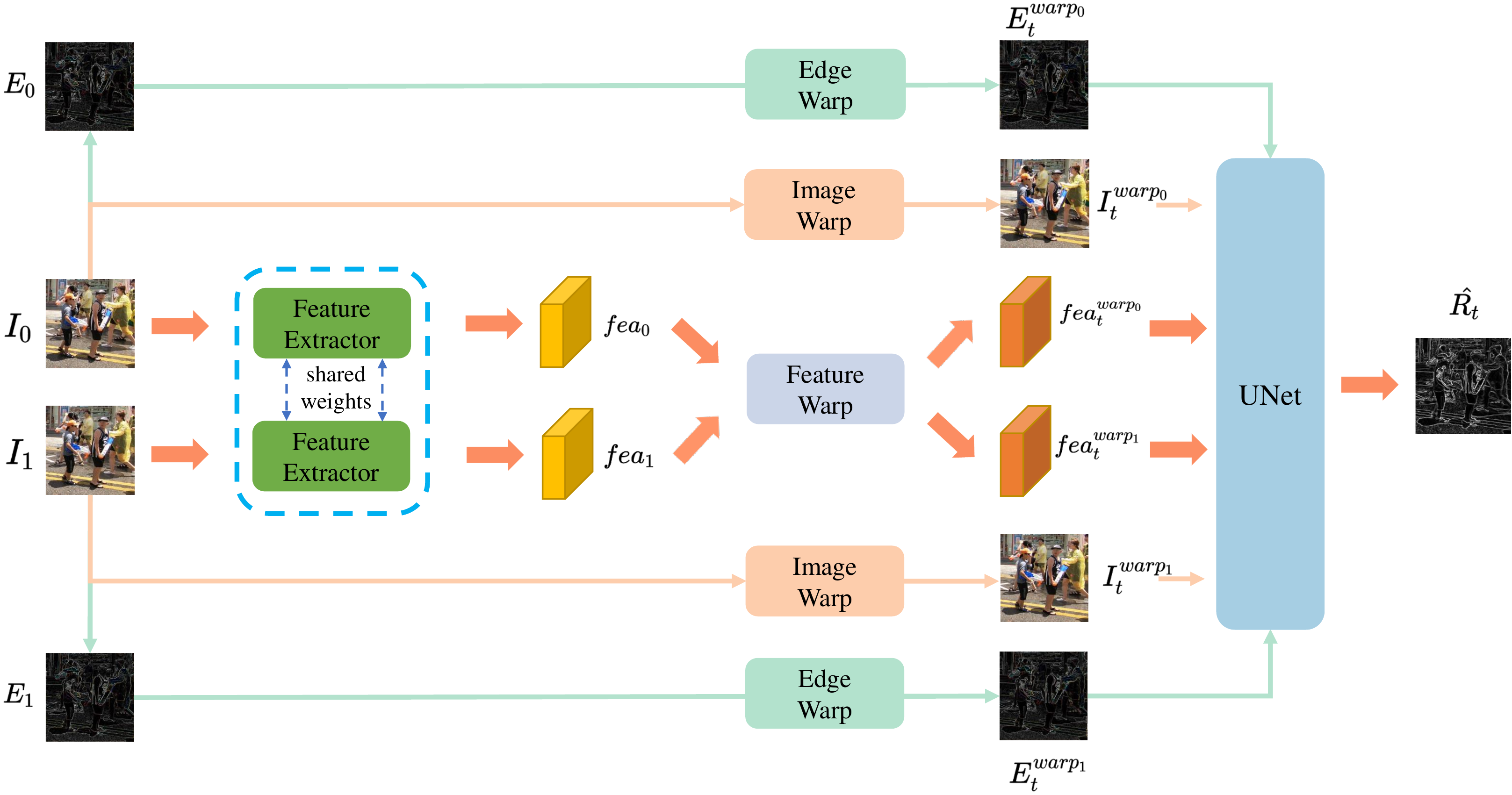}
    \\
    \caption{
        \textbf{XPixel: The residual contextual synthesis network.}
    }
    \label{fig:synthesis_xpixel}
\end{figure}
\begin{figure}[t!]
    \centering
    \includegraphics[width=\linewidth]{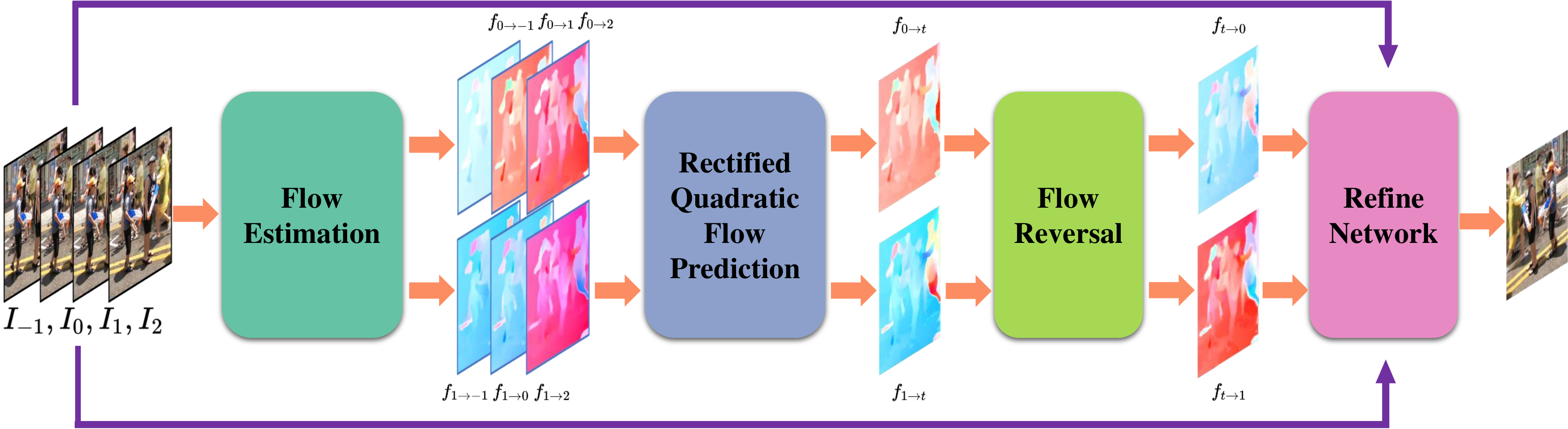}
    \\
    \caption{
        \textbf{XPixel: The rectified quadratic flow prediction.}
    }
    \label{fig:qvi_xpixel}
\end{figure}

\Paragraph{Residual contextual synthesis network.}
Inspired by \cite{Niklaus_2018_CVPR}, a residual contextual synthesis network~(RCSN) is designed to warp and exploit the contextual information in high-dimension feature space.
Specifically, the $conv1$ layer of the pre-trained ResNet18~\cite{He_2016_CVPR} is adopted to capture the contextual information of input frame $I_0$ and $I_1$.
Then we apply back warping on the features with $f_{0 \rightarrow t}$ to attain pre-warped features.
The edge information of $I_0$ and $I_1$ is also extracted and warped to preserve and leverage more structural information.
For simplicity, we calculate the gradient of each channel of the input frame as edge information.
Afterward, we feed the warped images, edges, and features into a small network to synthesize a residual map $\hat{R_t}$.
The refined output is obtained by $\hat{I}_t^{refined}=\hat{I}_t+\hat{R}_t$.
Fig.~\ref{fig:synthesis_xpixel} shows the overall organization of the network.

\Paragraph{Rectified quadratic flow prediction.}
Given four input consecutive frames $I_{-1}$, $I_{0}$, $I_{1}$ and $I_{2}$, the goal of video temporal super-resolution is to interpolate an intermediate frame $I_{t}$, where $t \in (0, 1)$.
Following the quadratic model~\cite{qvi_nips19}, a least square estimation method is utilized to improve the accuracy of quadratic frame interpolation further.
Unlike the original QVI~\cite{qvi_nips19}, all four frames (or three equations) are used for the quadratic model. 
If the motion of the input four frames basically conforms to the uniform acceleration model, the following equations should hold or approximately hold:
\begin{equation}
    \begin{split}
        f_{0 \rightarrow -1} &= -v_0 + 0.5 a, \\
        f_{0 \rightarrow 1} &= v_0 + 0.5 a, \\
        f_{0 \rightarrow 2} &= 2v_0 + 2a,
    \end{split}
\end{equation}
where $f_{0\rightarrow t}$ denotes the displacement of the pixel from frame 0 to frame $t$, $v_0$ represents the velocity at frame 0, and $a$ is the acceleration of the quadratic motion model.
The equations above can be transformed into a matrix form:
\begin{equation}
    \underbrace{\begin{bmatrix}
        -1 & 0.5 \\
         1 & 0.5 \\
         2 & 2 \\
    \end{bmatrix}}_{A}
    \begin{bmatrix}
        v_0 \\
        a \\
    \end{bmatrix}
    =
    \underbrace{\begin{bmatrix}
        f_{0 \rightarrow -1} \\
        f_{0 \rightarrow 1} \\
        f_{0 \rightarrow 2} \\
    \end{bmatrix}}_{b},
\end{equation}
where the solution $x^*= [v_0^* \quad a^*]^T$ can be derived as $x^* = [A^TA]^{-1}A^Tb$.
Then, the intermediate flow could be formulated as $f_{0\rightarrow t}=v_0^*t+ \frac{1}{2} a^*t^2$.
When the motion of input four frames approximately fits the quadratic assumption, the LSE solution can make a better estimation of $f_{0\rightarrow t}$.
As the real scenes are usually complex, several simple rules are adopted to discriminate whether the motion satisfies the quadratic assumption.
For any pair picked from $I_{-1}$, $I_{1}$ and $I_{2}$, an acceleration $a_i$ is calculated as:
\begin{equation}
    \begin{split}
        a_1 &= f_{0 \rightarrow -1} + f_{0 \rightarrow 1} \\
        a_2 &= \frac{2}{3}f_{0 \rightarrow -1} + \frac{1}{3}f_{0 \rightarrow 2} \\
        a_3 &= f_{0 \rightarrow 2} + 2f_{0 \rightarrow 2}
    \end{split}
\end{equation}

Theoretically, for quadratic motion, $a_1$, $a_2$ and $a_3$ should be in the same direction and approximately equal to each other.
If the orientation of those accelerations is not consistent, the model will be directly degenerated to the original QVI; otherwise, we adopt the following weight function to fuse the rectified flow and the original quadratic flow according to the proximity of quadratic model:
\begin{equation}
    \alpha(z) = -\frac{1}{2}\left[ \frac{e^{\omega(z-\gamma)}-e^{-\omega(z-\gamma)}}{e^{\omega(z-\gamma)}+e^{-\omega(z-\gamma)}} \right]+\frac{1}{2},
\end{equation}
where $z=|a_1-a_2|$, $\omega$ is the axis of symmetry, and $\gamma$ is the stretching factor.
Empirically, $\omega$ and $\gamma$ are set to 5 and 1, respectively.
The final $v_0$ and $a$ is obtained by:
\begin{equation}
    \begin{split}
        v_0 &= \alpha* v_0^{lse} + (1-\alpha)* v_0^{ori}, \\
        a &= \alpha* a^{lse} + (1-\alpha)* a^{ori},
    \end{split}
\end{equation}
where $^{lse}$ and $^{ori}$ represent the rectified LSE prediction and original prediction, respectively.
Fig.~\ref{fig:qvi_xpixel} demonstrates how the rectified quadratic flow is estimated.

\begin{figure}[t!]
    \centering
    \includegraphics[width=\linewidth]{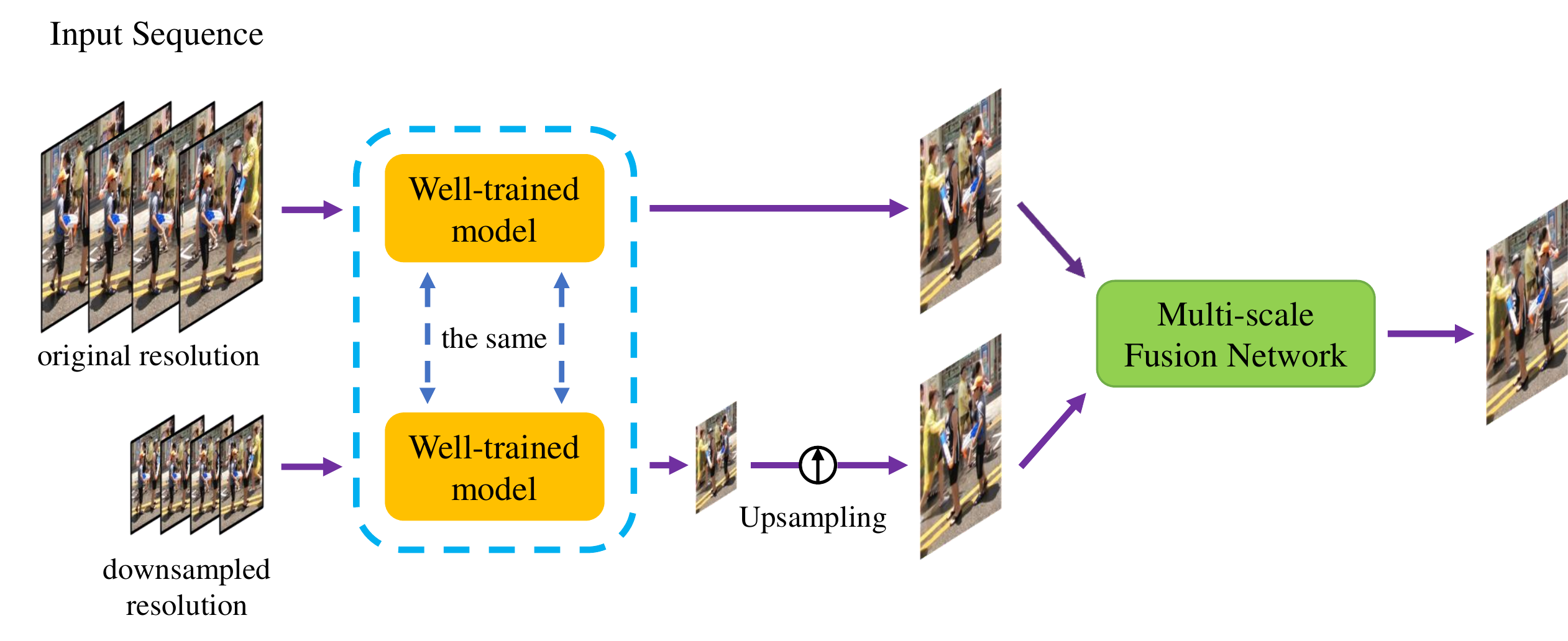}
    \\
    \caption{
        \textbf{XPixel: The multi-scale fusion network.}
    }
    \label{fig:fusion_xpixel}
\end{figure}

\Paragraph{Multi-scale fusion network.}
Finally, a novel multi-scale fusion network is proposed to capture various motions at a different level and further boost the performance.
This fusion network can be regarded as a learnable augmentation process at different resolutions.
Once we obtain a well-trained interpolation model from previous stages, we feed the model with one input sequence and its downsampled counterpart.
Then a fusion network is trained to predict a pixel-wise weighted map $M$ to fuse the results.
\begin{equation}
    M = F(Q(I_{in}), Up(Q(Down(I_{in})))),
\end{equation}
where $Q$ represents the well-trained QVI model, $Down(.)$ and $Up(.)$ denote the downsampling and upsampling operations, respectively.
Then, the final output interpolated frame is formulated as follows, as illustrated in Fig.~\ref{fig:fusion_xpixel}:
\begin{equation}
    \hat{I}_t^{final}=M*Q(I_{in}) + (1-M)*Up(Q({Down(I_{in})})).
\end{equation}
%
%
The proposed method is implemented under Python 3.6 and PyTorch 1.2 environments.
The training requires about 5 days using 4 RTX 2080 Ti GPUs.

\begin{figure}[t!]
    \centering
    \includegraphics[width=\linewidth]{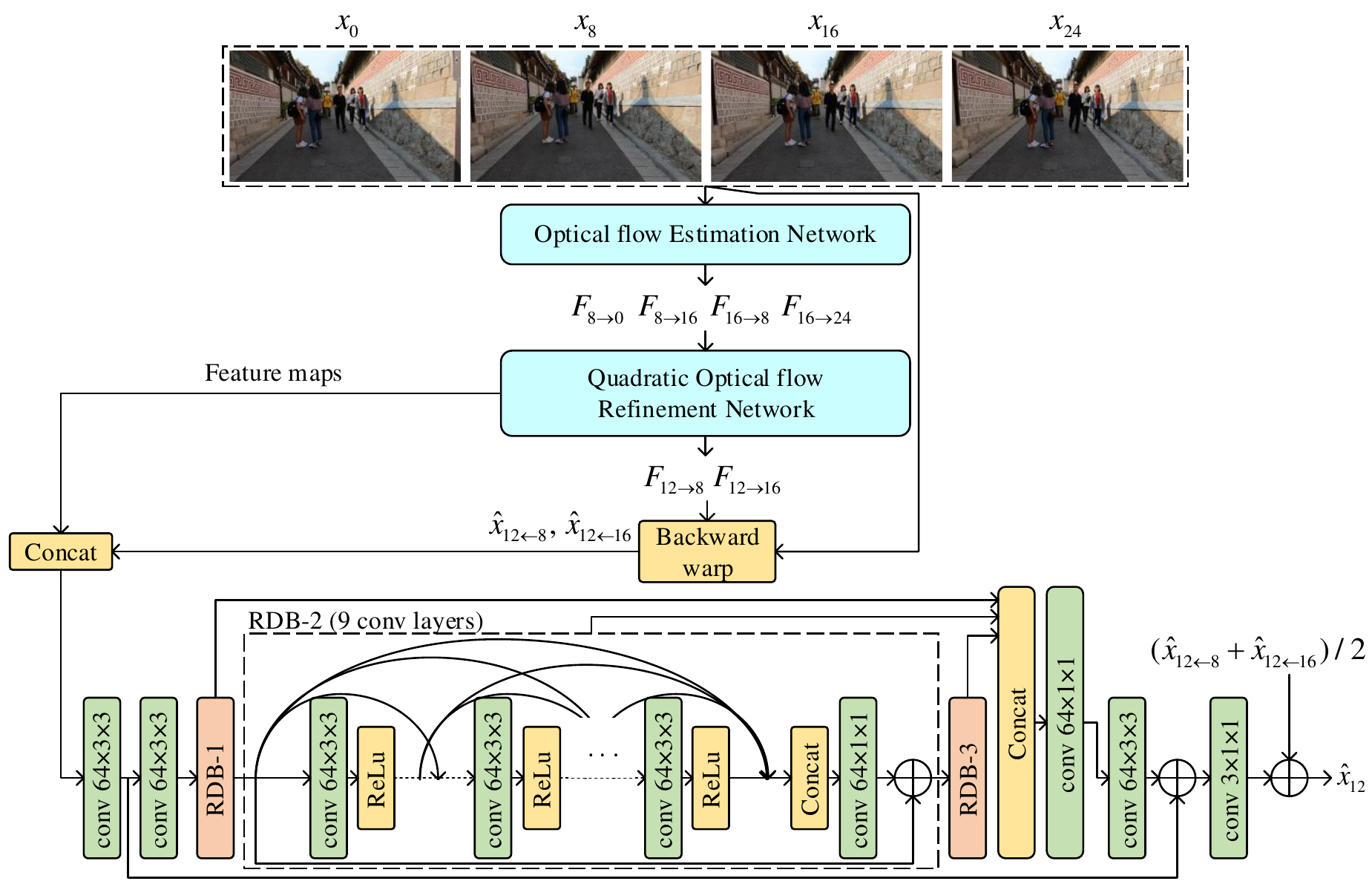}
    \caption{
        \textbf{KAIST-VICLAB: The architecture of the multi-frame synthesis network.}
        The optical flow estimation network is composed of the PWC-net~\cite{sun2018pwc}, and the quadratic optical flow refinement network is composed of the U-Net model similar to the original QVI method~\cite{qvi_nips19}.
    }
    \label{fig:model_kaist}
\end{figure}

\subsection{KAIST-VICLAB}
\Paragraph{Method.}
KAIST-VICLAB team proposes the quadratic video frame interpolation with a multi-frame synthesis network.
%
%
%
Inspired by the quadratic video interpolation~\cite{qvi_nips19} and the context-aware synthesis for video interpolation~\cite{Niklaus_2018_CVPR}, the proposed approach leverages these types of flow-based method.
The REDS\_VTSR dataset~\cite{Nah_2019_ICCV_Workshops_VTSR} is composed of high-resolution frames with much more complex motion than typical frame interpolation dataset~\cite{xue2019video}.
Since the optical flow is vulnerable to complex motion or occlusion, the proposed method attempts to handle the vulnerability by a nonlinear synthesis method with more information from neighbor frames.
Therefore, the proposed algorithm first estimates the optical flows between the intermediate frame and the neighboring frames using four adjacent frames.
The method for estimating the optical flows follows the same method for estimating the optical flows of the intermediate frame by a quadratic method based on the PWC-net~\cite{sun2018pwc} in~\cite{qvi_nips19}.
Unlike the QVI~\cite{qvi_nips19} method, the proposed method utilizes all the four neighboring frames for the synthesis network and incorporates a nonlinear synthesis network into the last part of our total network to better adapt to the frames containing complex motion or occlusion.

\Paragraph{Details.}
The proposed model is composed of the residual dense network~\cite{zhang2018residual}, and the inputs of the synthesis network are not only the feature maps obtained by estimating the optical flows using four neighboring frames, but also the warped neighboring frames.
Also, the output of the synthesis network is the corresponding intermediate frame.
Motivated by TOFlow~\cite{xue2019video}, the PWC-net for the optical flow estimation is fine-tuned on the REDS\_VTSR dataset.
More details about the proposed network are illustrated in Fig.~\ref{fig:model_kaist}.
The intermediate frames are optimized using the Laplacian loss~\cite{Niklaus_2018_CVPR} on three different scales and VGG loss.
%
%
%
The method is developed under Python 3.6 and PyTorch 1.4 environments, using one TITAN Xp GPU.
The training takes about 3 days.

\begin{figure}[t!]
    \centering
    \includegraphics[width=\linewidth]{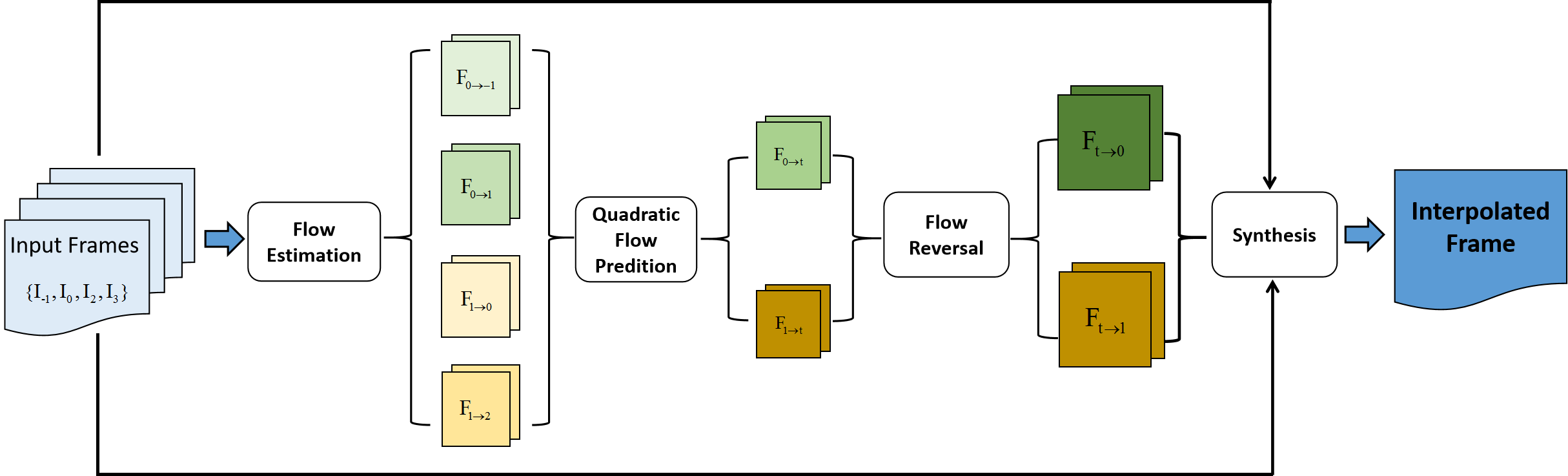}
    \\
    \caption{
        \textbf{BOE-IOT-AIBD: The architecture of the MSQI model.}
    }
    \label{fig:model_boe}
\end{figure}

\subsection{BOE-IOT-AIBD}
\Paragraph{Method.}
BOE-IOT-AIBD team has proposed the Multi Scale Quadratic Interpolation (MSQI) approach for the VTSR task.
The model is trained on the REDS\_VTSR dataset~\cite{Nah_2019_ICCV_Workshops_VTSR} of three different scales.
At each scale, the proposed MSQI employs PWC-Net~\cite{sun2018pwc} to extract optical flows and further refines the flow continuously through three modules: quadratic acceleration, flow reverse, and U-Net~\cite{ronneberger2015u} refine module.
Finally, a synthesis module interpolates the output frame by warping in-between inputs and refined flows.
During the inference, a post-processing method is applied to grind the interpolated frame.

\Paragraph{Details.}
Fig.~\ref{fig:model_boe} illustrates an overview of the proposed MSQI method.
Firstly, the MSQI model has trained on sequences from the REDS\_VTSR dataset~\cite{Nah_2019_ICCV_Workshops_VTSR}, using $\times 4$ lower resolution~($320 \times 180$) and frame rate~(15fps) for quick convergence.
Then, the model is fine-tuned on sequences of higher resolution~($640 \times 360$).
In the last stage, the MSQI model is trained using full-size images~($1280 \times 720$) with a 15fps frame rate.
During the inference time, a smoothing function is adopted for post-processing.
Also, quantization noise is injected during the training and inference phase.
The proposed method also adds the quantization noise to input frames at the inference phase and employs a smooth function to grind the interpolated frame.
The method adopts the pre-trained off-the-shelf optical flow algorithm, PWC-Net~\cite{sun2018pwc}, from the SenseSlowMo~\cite{qvi_iccvw19,qvi_nips19} model.
Compared to the baseline SenseSlowMo~\cite{qvi_iccvw19,qvi_nips19} architecture, the proposed MSQI converges better and focuses on multiple resolutions.
Python 3.7 and PyTorch environments are used on a V100 GPU server with a total of 128GB VRAM.

%

\subsection{TTI}
\Paragraph{Method.}
TTI team has proposed a temporal super-resolution method that copes large motions by reducing an input sequence's spatial resolution.
The approach is inspired by STARnet~\cite{haris2020space} model.
With the idea that space and time are related, STARnet jointly optimizes three tasks, i.e., spatial, temporal, and spatio-temporal super-resolution.

\begin{figure}
    \centering
    \includegraphics[width=\linewidth]{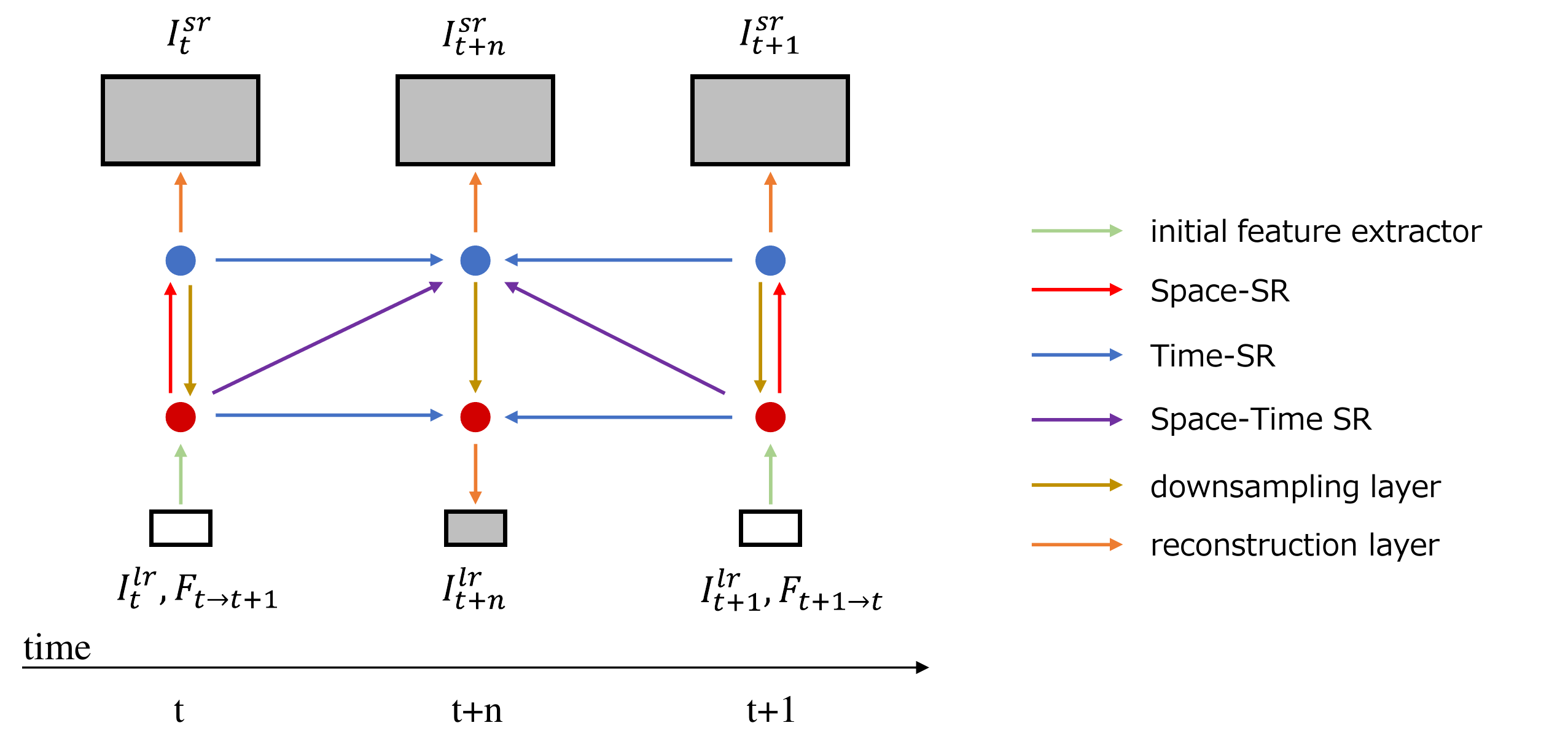}
    \\
    \caption{
        \textbf{TTI: The architecture of the STARnet model.}
        White and gray rectangles indicate input and output frames, respectively.
        %
        %
        For temporal super-resolution, two given frames are regarded as HR frames and downscaled to $I^{lr}_{t}$ and $I^{lr}_{t+1}$.
        From $I^{lr}_{t}$ and $I^{lr}_{t+1}$, the proposed STARnet acquires $I^{sr}_{t+n}$ as an interpolated frame in the original HR resolution.
    }
    \label{fig:model_tti}
\end{figure}

\Paragraph{Details.}
Fig.~\ref{fig:model_tti} illustrates the concept of the STARnet.
The model takes two LR frames $I^{lr}_t$  and $I^{lr}_{t+1}$, with bidirectional dense motion flows of these frames $F_{t\rightarrow t+1}$ and $F_{t+1\rightarrow t}$ as input.
The output consists of an in-between LR $I^{lr}_{t+n}$ and three SR frames $I^{sr}_t$, $I^{sr}_{t+n}$, and $I^{sr}_{t+1}$, where $n \in [0, 1]$ denotes the temporal interpolation rate.
While the STARnet model can be fine-tuned only for temporal super-resolution as proposed in~\cite{haris2020space}, the TTI team has employed the original spatio-temporal framework.
This is because large motions observed in the REDS~\cite{Nah_2019_CVPR_Workshops_REDS} dataset make it difficult to estimate accurate optical flows, which take an essential role for the VTSR task.
Therefore, the optical flows on the LR domain might support VTSR of HR frames in the REDS sequences.
In the proposed strategy, two given frames are resized to LR frames $I^{lr}_t$  and $I^{lr}_{t+1}$, and then fed into the network shown in Fig.~\ref{fig:model_tti} in order to acquire $I^{sr}_{t+n}$ in the original resolution.
For $\times 4$ temporal super-resolution, $n$ is set to $[0.25, 0.5, 0.75]$.
To realize those three temporal interpolation rates with one network, input flow maps $F_{t\rightarrow t+1}$ and $F_{t+1\rightarrow t}$ are scaled according to $n$ as follows:
\begin{equation}
    \begin{split}
        \hat{F}_{t\rightarrow t+1} &= n F_{t\rightarrow t+1}, \\ 
        \hat{F}_{t+1\rightarrow t} &= (1-n) F_{t+1\rightarrow t},
    \end{split}
\end{equation}
where $\hat{F}_{t\rightarrow t+1}$ and $\hat{F}_{t+1 \rightarrow t}$ denote the re-scaled flow maps that are used as input, respectively.
For higher accuracy, two changes are made to the original STARnet model.
First, a deformable convolution~\cite{dai2017deformable} is added at the end of T- and ST-SR networks to explicitly deal with object motions.
Second, the PWC-Net~\cite{sun2018pwc} is used to get flow maps, which show better performance than the baseline.

During training, each input frame is downscaled by half using bicubic interpolation.
Those frames are super-resolved to its original resolution by the S-SR network. 
The $\times 2$ S-SR network, which is based on the DBPN~\cite{haris2018deep} and RBPN~\cite{haris2019recurrent} models, is pre-trained on the Vimeo-90k dataset~\cite{xue2019video}.
Various data augmentations, such as rotation, flipping, and random cropping, are applied to input and target pairs.
All experiments are conducted using Python 3.6 and PyTorch 1.4 on Tesla V100 GPUs.

%

\section{Conclusion}
In the AIM 2020 VTSR challenge, 5 teams competed to develop state-of-the-art VTSR methods with the REDS\_VTSR dataset.
Top 3 methods leverage the quadratic motion modeling~\cite{qvi_nips19,qvi_iccvw19}, demonstrating the importance of accurate motion prediction. 
The winning team XPixel proposes Enhanced Quadratic Video Interpolation framework, which improves the QVI~\cite{qvi_nips19,qvi_iccvw19} method with three novel components.
Compared to AIM 2019 VTSR challenge~\cite{Nah_2019_ICCV_Workshops_VTSR}, there has been a significant PSNR improvement of 0.22dB on the 15 $\rightarrow$ 60 fps task over the QVI method~\cite{qvi_iccvw19}.
We also compare the submitted methods in a unified framework and provide a detailed analysis with specific example cases.
In our future challenge, we will encourage participants to develop 1) More efficient algorithms, 2) Perceptual interpolation methods, 3) Robust models on challenging real-world inputs.

\section*{Acknowledgments}
\label{sec:acknowledgments}
We thank all AIM 2020 sponsors: Huawei Technologies Co. Ltd., MediaTek Inc., NVIDIA Corp., Qualcomm Inc., Google, LLC and CVL, ETH Z{\"u}rich.

\appendix
\section{Teams and affiliations}
\label{sec:appendix}

\subsection*{AIM 2020 team}
\noindent\textit{\textbf{Title: }} AIM 2020 Challenge on Video Temporal Super-Resolution \\
\noindent\textit{\textbf{Members: }}
    \underline{Sanghyun Son$^1$ (sonsang35@gmail.com)},
    Jaerin Lee$^1$,
    Seungjun Nah$^1$,
    Radu Timofte$^2$,
    Kyoung Mu Lee$^1$\\
\noindent\textit{\textbf{Affiliations: }} \\
$^1$ Department of ECE, ASRI, Seoul National University~(SNU), Korea \\
$^2$ Computer Vision Lab, ETH Z{\"u}rich, Switzerland \\

\subsection*{XPixel}
\noindent\textit{\textbf{Title: }} Enhanced Quadratic Video Interpolation \\
\noindent\textit{\textbf{Members: }}
    \underline{Yihao Liu$^{1, 2}$ (liuyihao14@mails.ucas.ac.cn)},
    Liangbin Xie$^{1, 2}$,
    Li Siyao$^3$,
    Wenxiu Sun$^3$,
    Yu Qiao$^1$,
    Chao Dong$^1$\\
\noindent\textit{\textbf{Affiliations: }} \\
$^1$ Shenzhen Institutes of Advanced Technology, CAS \\
$^2$ University of Chinese Academy of Sciences \\
$^3$ SenseTime Research \\

\subsection*{KAIST-VICLAB}
\noindent\textit{\textbf{Title: }} Quadratic Video Frame Interpolation with Multi-frame Synthesis Network \\
\noindent\textit{\textbf{Members: }}
    \underline{Woonsung Park$^1$ (pys5309@kaist.ac.kr)},
    Wonyong Seo$^1$ (wyong0122@kaist.ac.kr),
    Munchurl Kim$^1$ (mkimee@kaist.ac.kr)\\
\noindent\textit{\textbf{Affiliations: }} \\
$^1$ Video and Image Computing Lab, Korea Advanced Institute of Science and Technology~(KAIST) \\

\subsection*{BOE-IOT-AIBD}
\noindent\textit{\textbf{Title: }} Multi Scale Quadratic Interpolation Method \\
\noindent\textit{\textbf{Members: }}
    \underline{Wenhao Zhang$^1$ (zhangwenhao@boe.com.cn)},
    Pablo Navarrete Michelini$^1$\\
\noindent\textit{\textbf{Affiliations: }} \\
$^1$ BOE, IOT AIBD, Multi Media Team \\

\subsection*{TTI}
\noindent\textit{\textbf{Title: }} STARnet for Video Frame Interpolation \\
\noindent\textit{\textbf{Members: }}
    \underline{Kazutoshi Akita$^1$ (sd19401@toyota-ti.ac.jp)},
    Norimichi Ukita$^1$\\
\noindent\textit{\textbf{Affiliations: }} \\
$^1$ Toyota Technological Institute~(TTI) \\

\bibliographystyle{splncs04}
\bibliography{egbib}
\end{document}